\documentclass[10pt,twocolumn,letterpaper]{article}

\usepackage{iccv}
\usepackage{times}
\usepackage{epsfig}
\usepackage{graphicx}
\usepackage{amsmath}
\usepackage{amssymb}
\usepackage{caption}
\usepackage{subcaption}
\usepackage{cite}

\usepackage{float}
\usepackage{multirow}
\usepackage{wrapfig}
\usepackage{array}
\usepackage{booktabs}
\usepackage{verbatim}
\usepackage{flushend}
\usepackage{marvosym}
\usepackage{wrapfig}
\usepackage[pagebackref=true,breaklinks=true,letterpaper=true,colorlinks,bookmarks=false]{hyperref}

\newcommand{\Paragraph}[1]{\vspace{1.25mm} \noindent \textbf{#1} \hspace{0mm}}

\iccvfinalcopy 

\def\iccvPaperID{722} 

\ificcvfinal\fi
\begin{document}

\title{$3$D Face Modeling From Diverse Raw Scan Data}

\author{Feng Liu, Luan Tran, Xiaoming Liu \\
Department of Computer Science and Engineering \\
Michigan State University, East Lansing MI 48824\\
{\tt \{liufeng6, tranluan, liuxm\}@msu.edu}
}

\maketitle
\ificcvfinal\thispagestyle{empty}\fi

\begin{abstract}
Traditional $3$D face models learn a latent representation of faces using linear subspaces from limited scans of a single database. The main roadblock of building a large-scale face model from diverse $3$D databases lies in the lack of dense correspondence among raw scans. To address these problems, this paper proposes an innovative framework to jointly learn a nonlinear face model from a diverse set of raw $3$D scan databases and establish dense point-to-point correspondence among their scans. Specifically, by treating input scans as unorganized point clouds, we explore the use of PointNet architectures for converting point clouds to identity and expression feature representations, from which the decoder networks recover their $3$D face shapes. Further, we propose a weakly supervised learning approach that does not require correspondence label for the scans. We demonstrate the superior dense correspondence and representation power of our proposed method, and its contribution to single-image $3$D face reconstruction.
\end{abstract}

\section{Introduction}

\begin{figure}[t]
\begin{center}
\includegraphics[width=0.99\linewidth]{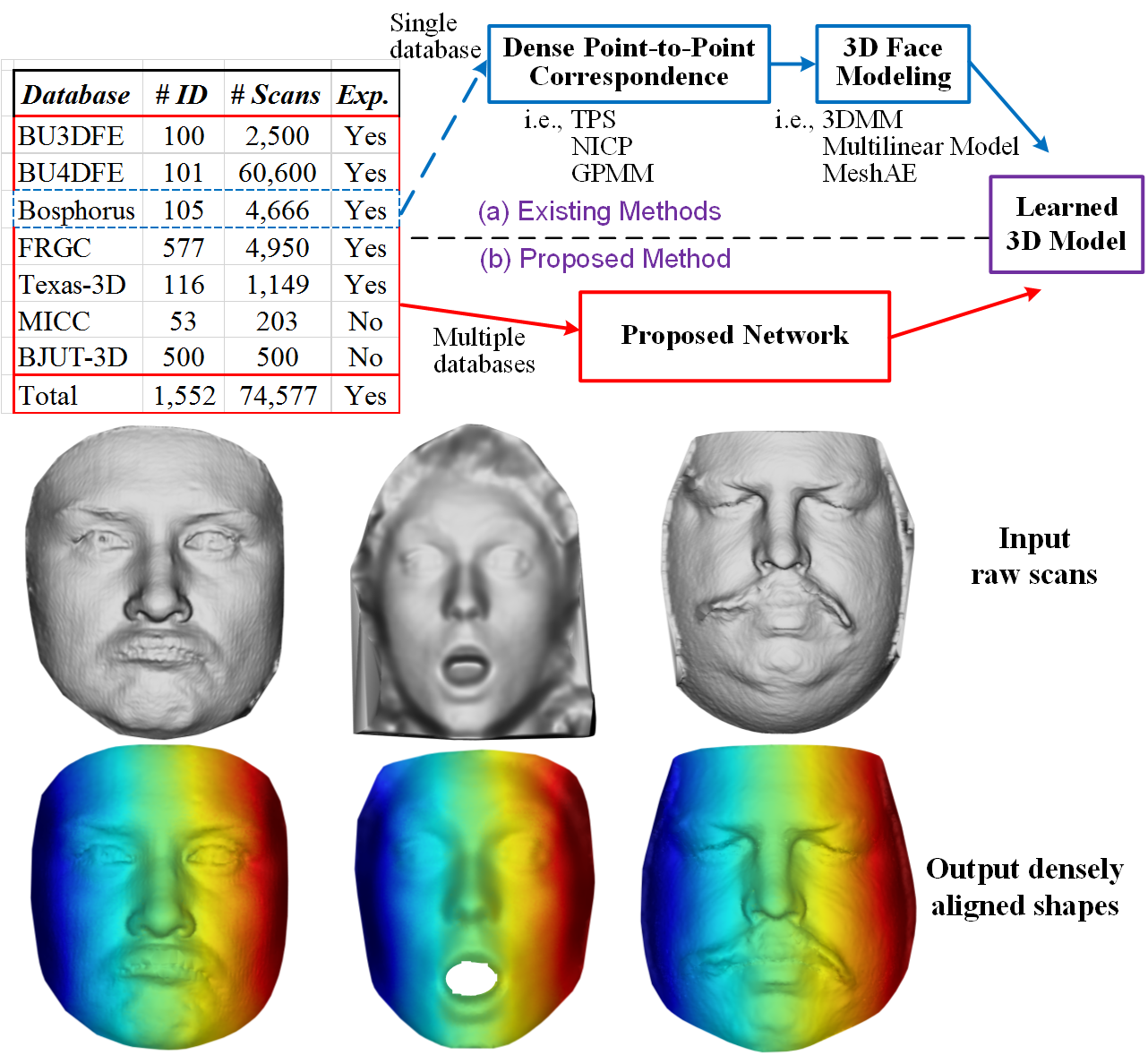}
\vspace{-2mm}
\end{center}
   \caption{\small Comparison between $3$D face modeling of (a) existing methods and (b) our proposed method. Dense point-to-point correspondence is a pre-requisite for the existing $3$D face modeling methods. Our proposed CNN-based approach learns face models \emph{directly} from raw scans of \emph{multiple} $3$D face databases and establish dense point-to-point correspondence among all scans (best viewed in color). Despite the diversity of scans in resolution and expression, our model can express the fine level of details.}
\label{fig:overview}
\vspace{-5mm}
\end{figure}

Robust and expressive $3$D face modeling is valuable for computer vision problems, \emph{e.g.}, $3$D reconstruction~\cite{blanz1999morphable, roth2016adaptive, gerig2018morphable, liu2016joint} and face recognition~\cite{taigman2014deepface, liu2018Disentangling, liu2018joint}, as well as computer graphics problems, \emph{e.g.}, character animation~\cite{cao2016real, han2017deepsketch2face}. The state-of-the-art $3$D face representations mostly adopt linear transformations~\cite{thies2015real, li2017learning, tewari2017mofa}, \emph{e.g.}, $3$D Morphable Model ($3$DMM) or higher-order tensor generalizations~\cite{vlasic2005face, cao2014facewarehouse, brunton2014multilinear, abrevaya2018multilinear}, \emph{e.g.}, Blendshapes Model. However, these linear models fall short of capturing the nonlinear deformations such as high-frequency details and extreme expressions. Recently, with the advent of deep learning, there have been several attempts at using deep neural networks for nonlinear data-driven face modeling~\cite{jiang2019disentangled, bagautdinov2018modeling, ranjan2018generating, on-learning-3d-face-morphable-model-from-in-the-wild-images}.

To model $3$D face shapes, a large amount of high-quality $3$D scans is required. 
The widely used $3$DMM-based BFM$2009$~\cite{paysan20093d} is built from scans of merely $200$ subjects in neutral expressions. Lack of expression may be compensated with expression bases from FaceWarehouse~\cite{cao2014facewarehouse} or BD-$3$FE~\cite{yin20063d}. After more than a decade, almost all existing models use less than $300$ training subjects. Such a small training set is far from adequate to describe the full variability of faces. Until recently, Booth \emph{et~al.}~\cite{Booth_2016_CVPR, booth2018large} build the first Large-Scale Face Model (LSFM) from neutral scans of $9{,}663$ subjects. Unfortunately, with only the resultant linear $3$DMM bases being released instead of the original scans, we cannot fully leverage this large database to explore different $3$D modeling techniques. 

In fact, there are many publicly available $3$D face databases, as shown in Fig.~\ref{fig:overview}. 
However, these databases are often used \emph{individually}, rather than \emph{jointly} to create large-scale face models.
The main hurdle lies in the challenge of estimating dense point-to-point correspondence for raw scans, which allows these scans to be organized in the same vector space, enabling analysis as a whole.

Dense point-to-point correspondence is one of the most fundamental problems in $3$D face modeling~\cite{gilani2018dense, fan2018dense}, which can be defined as in~\cite{fan2018dense}: given two $3$D faces $S$ and ${S}'$, the correspondence should satisfy three perspectives: i) $S$ and ${S}'$ have the same number of vertices; ii) the corresponding points share the same semantic meaning; iii) the corresponding points lie in the same local topological triangle context. 
Prior dense correspondence methods~\cite{blanz1999morphable, patel20093d, amberg2007optimal, gerig2018morphable}  lack either accuracy, robustness or automation. Moreover, few of them have shown success on {\it multiple} databases.
Beyond of the data scale, the challenge of dense correspondence for {\it multiple} databases is certainly escalated over single database: the quality of scans is often inevitably corrupted with artifacts (\emph{e.g.}, hair and eyebrows), missing data and outliers; facial morphology varies significantly due to expressions like mouth opening and closing; different databases contain high variability on the resolution.


To address these challenges, we propose a novel encoder-decoder to learn face models directly from raw $3$D scans of multiple diverse databases, as well as establish dense correspondence among them. 
Our approach provides: i) a PointNet-based encoder that learns nonlinear identity and expression latent representations of $3$D faces; ii) a corresponding decoder capable of establishing dense correspondence for scans with a variety of expressions and resolutions; iii) the decoder can be plugged into existing image-based encoders for $3$D face reconstruction. 
Specifically, by treating raw scans as unorganized point clouds, we explore the use of PointNet~\cite{qi2017pointnet} for converting point clouds to identity and expression representations, from which the decoder recovers their $3$D face shapes.


However, full supervision is often not available due to the lack of {\it ground-truth} dense correspondence. 
Thus, we propose a weakly-supervised approach with a mixture of synthetic and real $3$D scans. Synthetic data with topological ground truth helps to learn a shape correspondence prior in a supervised fashion, which allows us to incorporate order-invariant loss functions, \emph{e.g.}, Chamfer distance~\cite{fan2017point}, for unsupervised training of real data. 
Meanwhile, a surface normal loss retains the original high-frequency details. 
For regularization, we use the edge length loss to encourage the triangulation topology on the template and the reconstructed point cloud to be the same. 
Finally, a Laplacian regularization loss improves the performance of mouth regions with extreme expressions. 
The above strategies allow the network to learn from a large set of raw $3$D scan databases without any label on correspondences. 
In summary, the contributions of this work include:

$\diamond$ We propose a new encoder-decoder framework that for the first time jointly learns face models directly from raw scans of multiple $3$D face databases and establishes dense correspondences among all scans. 

$\diamond$ We devise a weakly-supervised learning approach and several effective loss functions for the proposed framework that can leverage known correspondences from synthetic data and relax the Chamfer distance loss for vertex correspondence in an \emph{unsupervised} fashion.

$\diamond$ We demonstrate the superiority of our nonlinear model in preserving high-frequency details of $3$D scans, providing compact latent representation, and applications of single-image $3$D face reconstruction.


\begin{table}[t!]
\renewcommand\arraystretch{0.98}
\newcommand{\tabincell}[2]{\begin{tabular}{@{}#1@{}}#2\end{tabular}}
\scriptsize
\centering
\caption{\small Comparison of $3$D face modeling from scans. 'Exp.' refers to whether learns the expression latent space, 'Corr.' refers to whether requires densely corresponded scans in training.}
\vspace{-2mm}
\begin{tabular}{l |c| c | c | c | c  c }
\toprule
Method &  Dataset & Lin./nonL. &\#Subj. & Exp. &  Corr.\\ 
\hline\hline
BFM~\cite{paysan20093d}                      & BFM  & Linear & $200$      & No   & Yes\\
GPMMs~\cite{luthi2018gaussian}               & BFM  & Linear& $200$      & Yes  & Yes\\
LSFM~\cite{Booth_2016_CVPR, booth2018large}  & LSFM    & Linear& $9{,}663$  & No   & Yes\\
LYHM~\cite{dai20173d}  & LYHM    & Linear& $1{,}212$  & No   & Yes\\
\hline
Multil.~model~\cite{cao2014facewarehouse}& FWH  & Linear& $150$      & Yes  &  Yes\\
FLAME~\cite{li2017learning}                  &  \tabincell{c}{CAESAR\\ D3DFACS} & Linear & \tabincell{c}{$3{,}800$\\ $10$}  & Yes  &  Yes\\
\hline
VAE~\cite{bagautdinov2018modeling}           &  Proprietary  & Nonlin. & $20$       & No   & Yes\\
MeshAE~\cite{ranjan2018generating}           & COMA & Nonlin. & $12$       & No   & Yes \\
Jiang~\etal~\cite{jiang2019disentangled}     & FWH  & Nonlin. & $150$      & Yes   & Yes \\
\hline
\textbf{Proposed}                                         &  \textbf{$7$ datasets}   & \textbf{Nonlin.} & $\mathbf{1,552}$    & \textbf{Yes}   & \textbf{No}\\
\bottomrule
\end{tabular}
\label{tab:3D_modeling_review}
\vspace{-4mm}
\end{table}

\section{Related Work}

\begin{figure*}[t]
\begin{center}
\includegraphics[width=0.771\linewidth]{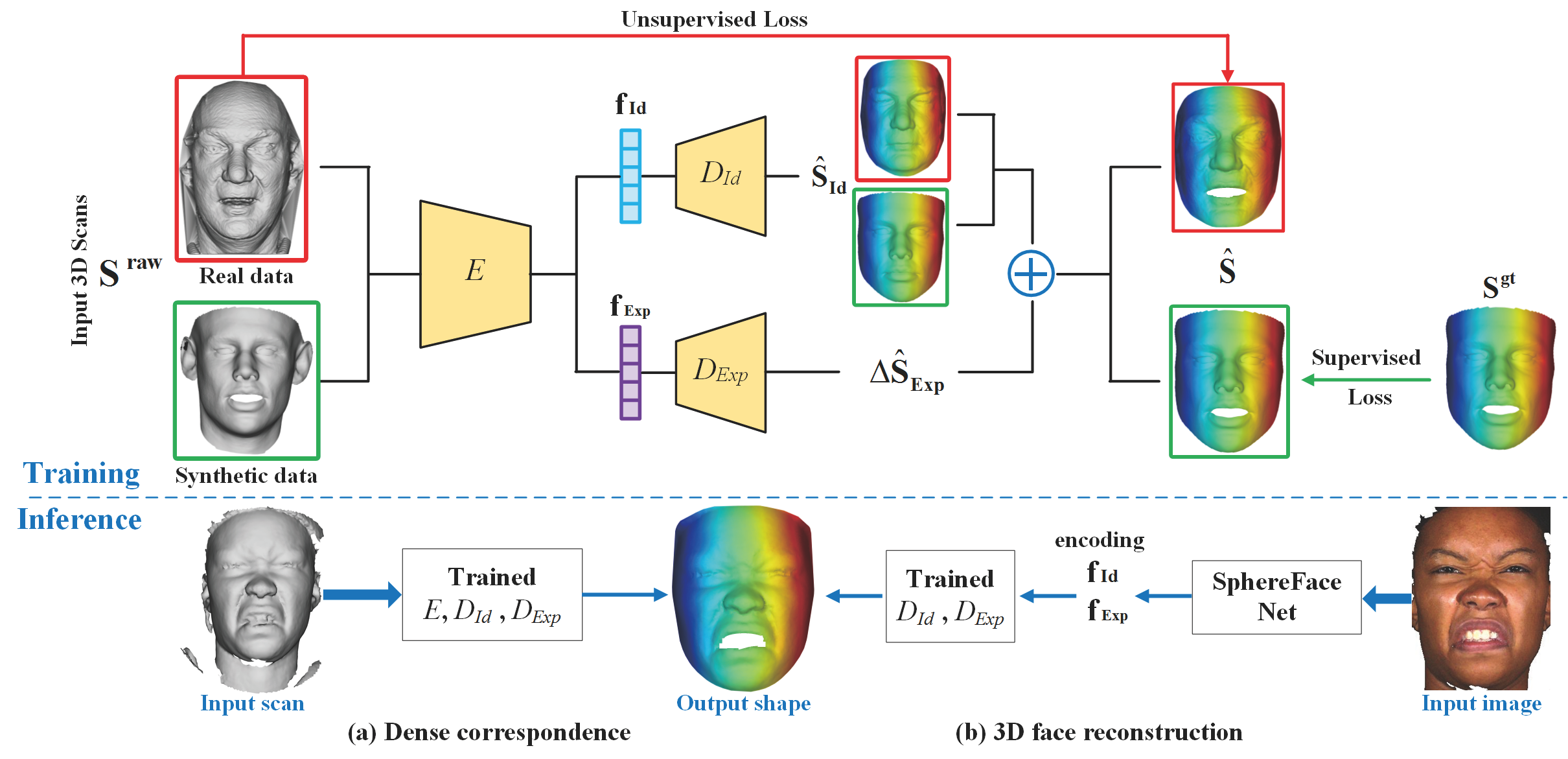}
\vspace{-3mm}
\end{center}
   \caption{\small Overview of our $3$D face modeling method. A mixture of synthetic and real data is used to train the encoder-decoder network with supervised (\textcolor{green}{green}) and unsupervised (\textcolor{red}{red}) loss. Our network can be used for $3$D dense correspondence and $3$D face reconstruction.}
\label{fig:flowchart}
\vspace{-3mm}
\end{figure*}

\textbf{$3$D Face Modeling.}
Traditional $3$DMMs~\cite{blanz1999morphable, blanz2003face} model geometry variation from limited data via PCA.
Paysan \emph{et~al.}~\cite{paysan20093d} build BFM$2009$, the publicly available morphable model in neutral expression, which is extended to emotive face shapes~\cite{amberg2008expression}. 
Gerig \emph{et al.}~\cite{luthi2018gaussian, gerig2018morphable} propose the Gaussian Process Morphable Models (GPMMs) and release a new BFM$2017$. 
Facial expressions can also be represented with higher-order generalizations. Vlasic \emph{et al.}~\cite{vlasic2005face} use a multilinear tensor-based model to jointly represent the identity and expression variations. FaceWarehouse (FWH)~\cite{cao2014facewarehouse} is a popular multilinear $3$D face model. The recent FLAME model~\cite{li2017learning} additionally models head rotation. 
However, all these works adopt a linear space, which is over-constrained and might not well represent high-frequency deformations. 

Deep models have been successfully used for $3$D face fitting, 
which recovers $3$D shape from $2$D images~\cite{zhu2016CVPR, tran2016regressing, richardson2016learning, Dou2017End, tewari2017mofa,jourabloo2017pose_iccv, jourabloo2016large, jourabloo2017poseijcv}. However, in these works the linear model is learned a-priori and fixed during fitting, unlike ours where the nonlinear model is learned during training. 


In contrast, applying CNN to learn more powerful $3$D face models has been largely overlooked. 
Recently, Tran \emph{et al.}~\cite{tran2018nonlinear,towards-high-fidelity-nonlinear-3d-face-morphable-model} learn to regress $3$DMM representation, along with the decoder-based models. 
SfSNet~\cite{sengupta2018sfsnet} learns shape, albedo and lighting decomposition of a face, from $2$D images, instead of $3$D scans.
Bagautdinov~\etal~\cite{bagautdinov2018modeling} learn nonlinear face geometry representations directly from UV maps via a VAE. 
Ranjan~\etal~\cite{ranjan2018generating} introduce a convolutional mesh autoencoder to learn nonlinear variations in shape and expression. 
Note that \cite{bagautdinov2018modeling,ranjan2018generating} train with no more than $20$ subjects and encode the $3$D data to a {\it single} latent vector. Jiang~\etal~\cite{jiang2019disentangled} extend 
~\cite{ranjan2018generating} to decompose a $3$D face into identity and expression parts. Unlike our work, these three methods require densely corresponded $3$D scans in training. 
We summarize the comparison in Tab.~\ref{tab:3D_modeling_review}.


\textbf{$3$D Face Dense Correspondence.}
As a fundamental shape analysis task, correspondence has been well studied in the literature. 
Shape correspondence, \emph{a.k.a.}~registration, alignment or simply matching~\cite{van2011survey}, finds a meaningful mapping between two surfaces. 
The granularity of mapping varies greatly, from semantic parts~\cite{groueix20183d, rodola2017partial}, group~\cite{cootes2004groupwise}, to points~\cite{kovnatsky2013coupled}.
Within this range, \textbf{point-to-point} correspondences for $3$D face is the most challenging and strict one. 
In original $3$DMM~\cite{blanz1999morphable}, the $3$D face dense correspondence is solved with a regularized form of optical flow as a cylindrical image registration task. 
This is only effective in constrained settings, where subjects share similar ethnicities and ages. 
To overcome this limitation, Patel and Smith~\cite{patel20093d} use a Thin Plate Splines (TPS)~\cite{bookstein1989principal} warp to register scans into a template. 
Alternatively, Amberg~\emph{et al.}~\cite{amberg2007optimal} propose an optimal step Nonrigid Iterative Closest Point (NICP) for registering $3$D shapes. 
Booth \emph{et al.}~\cite{Booth_2016_CVPR, booth2018large} quantitatively compare these three popular dense correspondence techniques in learning $3$DMM. 
Additional extensions are also proposed~\cite{zhang2016functional,gilani2017deep,gerig2018morphable,fan2018dense}.


Many algorithms~\cite{grewe2016fully, bolkart2015groupwise, abrevaya2018multilinear} treat dense correspondence as a $3$D-to-$3$D model fitting problem. 
E.g., \cite{bolkart2015groupwise} propose a multilinear groupwise model for $3$D face correspondence to decouple identity and expression variations. Abrevayaemph~\emph{et~al.}~\cite{abrevaya2018multilinear} propose a $3$D face autoencoder with a CNN-based depth image encoder and multilinear model as a decoder for $3$D face fitting.
However, these methods require $3$D faces with an initial correspondence as input and the correspondence problem is considered in the restrictive space expressed by the model. Although insightful and useful, a chicken-and-egg problem still remains unsolved~\cite{fan2018dense}. 

To summarize, prior work tackle the problems of $3$D face modeling, and $3$D face dense correspondence \emph{separately}. 
However, dense correspondence is a prerequisite for modeling. 
If the correspondence has errors, they will accumulate and propagate to $3$D modeling. 
Therefore, these two problems are highly relevant and our framework for the first time tackles them  simultaneously.

%

\section{Proposed Method}
This section first introduces a composite $3$D face shape model with latent representations. We then present the mixture training data and our encoder-decoder network. We finally provide implementation details and face reconstruction inference. Figure~\ref{fig:flowchart} depicts the overview of our method.

\subsection{Problem Formulation}
In this paper, the output $3$D face scans are represented as point clouds. Each \emph{densely aligned} $3$D face $\mathbf{S}\in\mathbb{R}^{n\times3}$ is represented by concatenating its $n$  vertex coordinates as, 
\begin{equation}
\mathbf{S} = [x_{1}, y_{1}, z_{1}; x_{2}, y_{2}, z_{2}; \cdots; x_{n}, y_{n}, z_{n}].
\end{equation}
We assume that a $3$D face shape is composed of identity and expression deformation parts,
 \begin{equation}
\mathbf{S} = \mathbf{S}_{Id} + \Delta\mathbf{S}_{Exp},
\end{equation}
where $\mathbf{S}_{Id}$ is the identity shape and $\Delta\mathbf{S}_{Exp}$ is expression difference. Since the identity and expression spaces are independent, we further assume these two parts can be described by respective latent representations, $\mathbf{f}_{Id}$ and $\mathbf{f}_{Exp}$. 

Specifically, as shown in Fig.~\ref{fig:flowchart}, we use two networks to decode shape component $ \mathbf{S}_{Id}$ and $\Delta \mathbf{S}_{Exp}$ from the corresponding latent representations. 
Formally, given a set of raw $3$D faces $\{\mathbf{S}^{raw}_{i}\}^{N}_{i=1}$, we learn an encoder $E:\mathbf{S}^{raw}{\rightarrow} \mathbf{f}_{Id}, \mathbf{f}_{Exp}$ that estimates the identity and expression shape parameters $\mathbf{f}_{Id}{\in} \mathbb{R}^{l_{Id}}, \mathbf{f}_{Exp}{\in} \mathbb{R}^{l_{Exp}}$, an identity shape decoder $D_{Id}: \mathbf{f}_{Id}{\rightarrow} \mathbf{S}_{Id}$, and an expression shape decoder $D_{Exp}: \mathbf{f}_{Exp}{\rightarrow} \mathbf{S}_{Exp}$ that decode the shape parameters to a $3$D shape estimation $\hat{\mathbf{S}}$.




Recent attempts to encode $3$D face shape in deep learning include point clouds, depth map~\cite{abrevaya2018multilinear}, UV map based mesh~\cite{bagautdinov2018modeling}, and mesh surface~\cite{ranjan2018generating, jiang2019disentangled}. Point clouds are a standard and popular $3$D face acquisition format used by Kinect, iPhone's face ID and structured light scanners. We thus design a deep encoder-decoder framework to directly consume unorganized point sets as input and output densely corresponded $3$D shapes. Before providing the algorithm details, we first introduce the real and synthetic training data served for the weakly-supervised learning. 


\begin{table}[t!]
\renewcommand\arraystretch{0.98}
\scriptsize
\centering
\caption{\small Summary of training data from related databases. }
\vspace{-2mm}
\begin{tabular}{l |c| c  c | c  c }
\toprule
Database &\#Subj. & \#Neu. &\#Sample & \#Exp. & \#Sample\\ 
\hline\hline
BU$3$DFE~\cite{yin20063d} &    $100$  &  $100$   & $1{,}000$ & $2{,}400$      & $2{,}400$\\
BU$4$DFE~\cite{yin2008high} &    $101$  &  $>$$101$  & $1{,}010$ & $>$$606$       & $2{,}424$\\
Bosphorus~\cite{savran2008bosphorus} & $105$  & $299$    & $1{,}495$ & $2{,}603$      & $2{,}603$\\
FRGC~\cite{phillips2005overview} & $577$  &$3{,}308$   & $6{,}616$ & $1{,}642$      & $1{,}642$\\
Texas-$3$D\cite{gupta2010texas}  & $116$  &$813$     & $1{,}626$ & $336$        & $336$\\
MICC\cite{bagdanov2011florence}      & $53$   &$103$     & $515$   & $\mathbf{-}$ & $\mathbf{-}$\\
BJUT-$3$D~\cite{baocai2009bjut}   & $500$  &$500$     & $5{,}000$ & $\mathbf{-}$ & $\mathbf{-}$\\
\hline
Real Data  &$1{,}552$   & $5{,}224$ & $17{,}262$& $7{,}587$      & $9{,}405$   \\      
\hline        
Synthetic Data &$1{,}500$ & $1{,}500$ &$15{,}000$&$9{,}000$&$9{,}000$ \\
\bottomrule
\end{tabular}
\label{tab:databases}
\vspace{-3mm}
\end{table}

\subsection{Training Data} 
\begin{figure}[t]
\begin{center}
\includegraphics[trim=20 0 12 0, clip, width=0.9\linewidth]{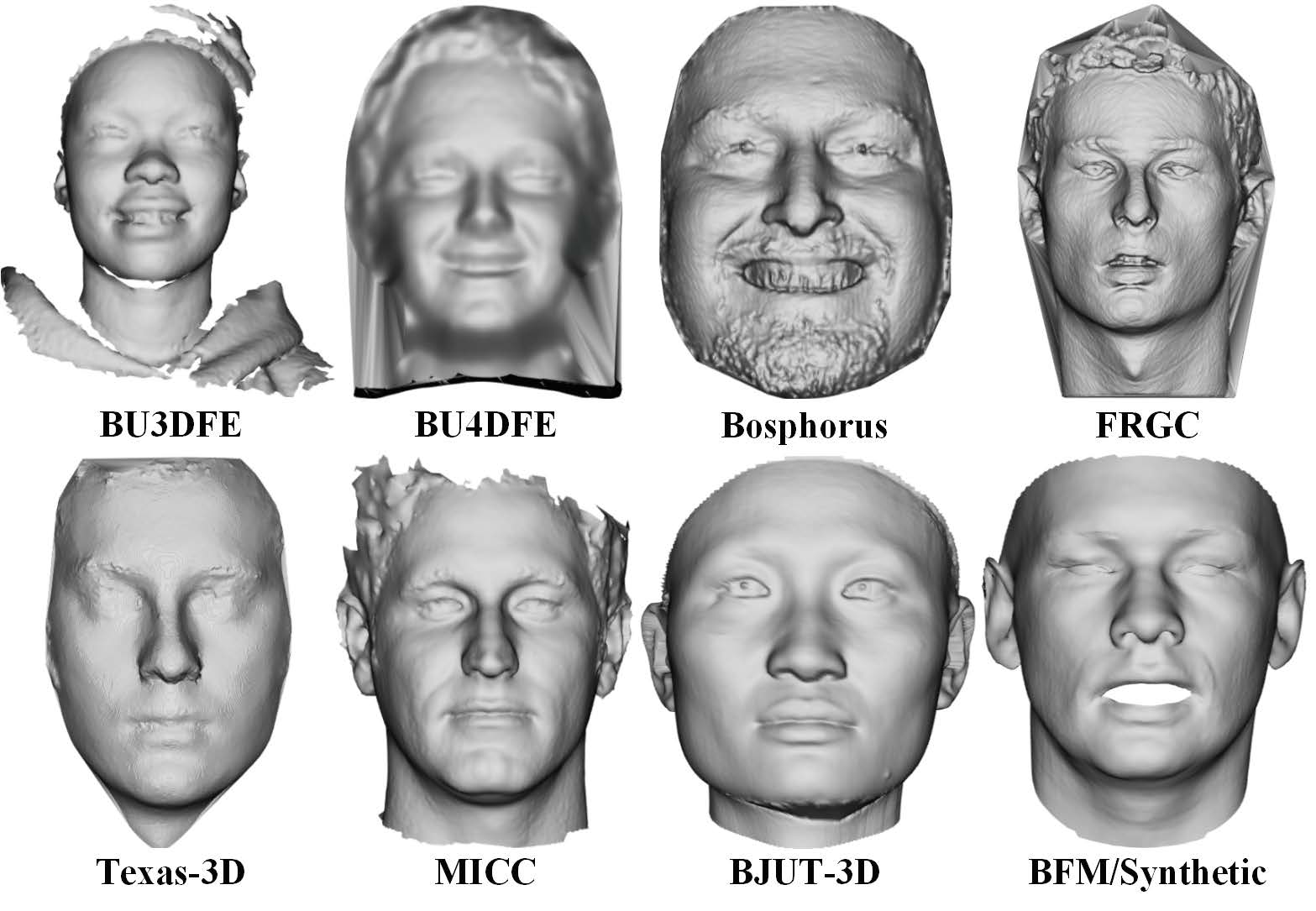}
\vspace{-3mm}
\end{center}
   \caption{\small One sample from each of eight $3$D face databases. They exhibit a wide variety of expressions and resolutions.}
\label{fig:databases}
\vspace{-3mm}
\end{figure}

To learn a robust and highly variable $3$D face model, we construct training data of seven \emph{publicly available} $3$D databases with a wide variety of identity, age, ethnicity, expression and resolution, listed in Tab.~\ref{tab:databases}. 
However, for these real scans, there are no associated ground-truth on dense correspondence. 
Recently, some $3$D databases are released such as $4$DFAB~\cite{cheng20184dfab}, Multi-Dim~\cite{liu2017multi} and UHDB $3$D~\cite{toderici2013uhdb11, wu2016rendering}. While including them may increase the amount of training data, they do not provide new types of variations beyond the seven databases. 
We do not use the occlusion and pose (self-occlusion) data of Bosphorus database, since extreme occlusion or missing data would break semantic correspondence consistency of $3$D faces.
For BU$4$DFE database, we manually select one neutral and $24$, expression scans per subject. 
To keep the balance between real and synthetic data, we use BFM$2009$ to  synthesize $3$D faces of $1{,}500$ subjects, and use $3$DDFA~\cite{zhu2015high} expression model to generate $6$ random expressions for each subject. 
Figure~\ref{fig:databases} shows one example scan from each of the eight databases.


\begin{figure}[t]
\begin{center}
\includegraphics[trim=0 39 0 0, clip, width=0.98\linewidth]{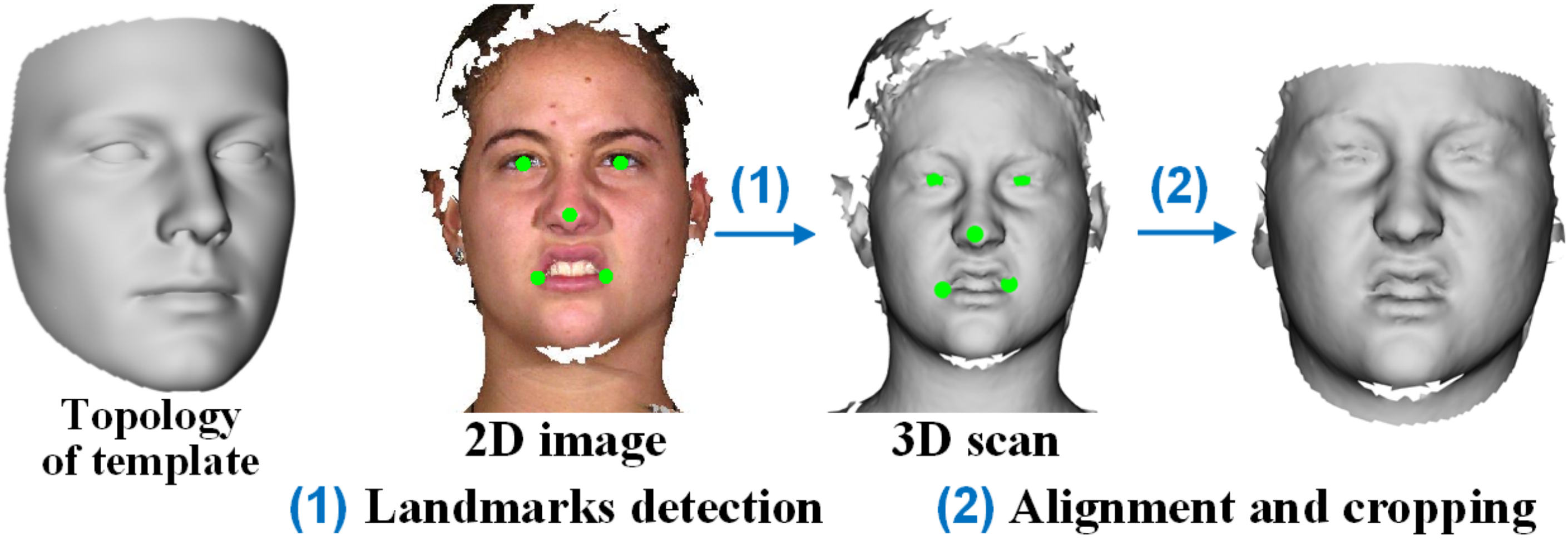}
\vspace{-3mm}
\end{center}
   \caption{\small Preprocessing. (1) Automatic $3$D landmark detection based on rendered images. (2) Guided by landmarks and predefined template, the $3$D scan is alignment and cropping.}
\label{fig:preprocessing}
\vspace{-4mm}
\end{figure}

\Paragraph{Preprocessing and data augmentation} 
As visualized in Fig.~\ref{fig:preprocessing}, we first predefine a template of $3$D face topology consisting of $n=29{,}495$ vertices and $58{,}366$ triangles, which is manually cropped from BFM mean shape. 
Then, we normalize the template into a unit sphere. 
The original synthetic examples contain $53{,}215$ vertices, after removing points on the tongue. 
For synthetic examples, we crop their face region with same topological triangulation as the template, perform the same normalization, and denote this resultant $3$D face set with ground-truth correspondence as $\{\mathbf{S}^{gt}_{i}\}_{i=1}^{M}$, whose number of vertices is also $n$.

Since raw scans are acquired from different distances, orientations or sensors, their point clouds exhibit enormous variations in pose and scale. 
Thus, before feeding them to our network, we apply a similarity transformation to align raw scans to the template by using five $3$D landmarks. 
Following~\cite{booth2018large}, we detect $2$D landmarks on the corresponding rendered images, from which we obtain $3$D landmarks by back-projection (Fig.~\ref{fig:preprocessing} (1)). 
After alignment, the points outside the unit sphere are removed. 
Finally, we randomly sample $n$ points as the input $\mathbf{S}^{input}_{i}\in \mathbb{R}^{n\times3}$. 
If the vertex number is less than $n$, we apply interpolating subdivision~\cite{kobbelt20003} before sampling. 
As in Tab.~\ref{tab:databases}, we perform data augmentation for neutral scans by repeating random sampling several times so that each subject has  $10$ neutral training scans. 
Note that the above preprocessing is also applied to synthetic data, except that their $3$D landmarks are provided by BFM. 
As a result, the point ordering of both input raw and synthetic data is \emph{random}. 

\subsection{Loss Function}
This encoder-decoder architecture is trained end-to-end. We define three kinds of losses to constrain the correspondence of the output shape and template, also to retain the original global and local information. The overall loss is: 
\begin{equation}
\mathcal{L} = \mathcal{L}^{vt}+\lambda_{1}\mathcal{L}^{normal}+\lambda_{2}\mathcal{L}^{edge},
\label{eqn_loss}
\end{equation}
where the vertex loss $\mathcal{L}^{vt}$ is to constrain the location of mesh vertices, normal loss $\mathcal{L}^{normal}$ is to enforce the consistency of surface normals, and edge length loss is to preserve the topology of $3$D faces.


Here, we consider two training scenarios: synthetic and real data. Supervision is typically available for the synthetic data with ground truth (supervised case), but real scans are obtained without correspondence label (unsupervised case).

\Paragraph{Supervised loss} In the supervised case, given the shape $\mathbf{S}^{gt}$ (and $\hat{\mathbf{S}}$) and predefined triangle topology, we can easily compute the corresponding surface normal $\mathbf{n}^{gt}$ (and $\hat{\mathbf{n}}$) and edge length $\mathbf{e}^{gt}$ (and  $\hat{\mathbf{e}}$). 
Therefore, for vertex loss, we can use $\mathcal{L}_{1}$ loss $\mathcal{L}^{vt}(\hat{\mathbf{S}}, \mathbf{S}^{gt}) {=} ||\mathbf{S}^{gt}-\hat{\mathbf{S}}||_{1}$. We measure the normal loss by cosine similarity distance $\mathcal{L}^{normal}(\hat{\mathbf{n}}, \mathbf{n}^{gt})=\frac{1}{n} \sum_{i}(1-\mathbf{n}^{gt}_{i} \cdot \hat{\mathbf{n}}_{i})$. If the predicted normal has a similar orientation as the ground truth, the dot-product $\mathbf{n}^{gt}_{i} \cdot \hat{\mathbf{n}}_{i}$ will be close to $1$ and the loss will be small, and vice versa. The third term $\mathcal{L}^{edge}$ encourages the ratio between edges length in the predicted shape and ground truth to be close to $1$. Following~\cite{groueix20183d}, edge length loss is defined as,
\begin{equation}
\mathcal{L}^{edge}(\hat{\mathbf{S}},\mathbf{S}^{gt})= \frac{1}{\#E} \sum_{(i,j)\in E }\left | \frac{\left \| \hat{\mathbf{S}}_{i} -\hat{\mathbf{S}}_{j}  \right \|}{\left \| \mathbf{S}^{gt}_{i} -\mathbf{S}^{gt}_{j} \right \|}-1 \right |,
\label{eqn_edge}
\end{equation}
where $E$ is the fixed edge graph of the template. 


\Paragraph{Unsupervised loss} In the case where the correspondences between the template and real scans are not available, we still optimize the reconstructions, but regularize the deformations toward correspondence. For reconstruction, we use the Chamfer distance as the $\mathcal{L}^{vt}(\hat{\mathbf{S}}, \mathbf{S}^{raw})$ between the input scans $\mathbf{S}^{raw}$ and the predicted $\hat{\mathbf{S}}$,
\begin{equation}
\begin{aligned}
 \mathcal{L}^{vt}(\hat{\mathbf{S}},\mathbf{S}^{raw})=\sum_{p\in\hat{\mathbf{S}}}\min_{q\in\mathbf{S}^{raw}}\left \| p - q \right \|_{2}^{2}+ \sum_{q\in\mathbf{S}^{raw}}\min_{p\in\hat{\mathbf{S}}}\left \| p-q \right \|_{2}^{2}, \\
 \end{aligned}
\end{equation}
where $p$ is a vertex in the predicted shape, $q$ is a vertex in the input scan. When $\min_{q\in\mathbf{S}^{raw}}\left \| p - q \right \|_{2}^{2}>\epsilon$ or $\min_{p\in\hat{\mathbf{S}}}\left \| p - q \right \|_{2}^{2}>\epsilon$, we treat $q$ as a flying vertex and the error will not be counted.

In this unsupervised case, we further define loss on the surface normal to characterize high-frequency properties, $\mathcal{L}^{normal}(\hat{\mathbf{n}}, \mathbf{n}^{raw}_{(q)})$, where $q$ is the closest vertex for $p$ that is found when calculating the Chamfer distance, and $\mathbf{n}^{raw}_{(q)}$ is the observed normal from the real scan. 
For the edge length loss, $\mathcal{L}^{edge}$  is defined the same as Eqn.~\ref{eqn_edge}. 
%

\begin{figure}[t]
\begin{center}
\includegraphics[width=0.8\linewidth]{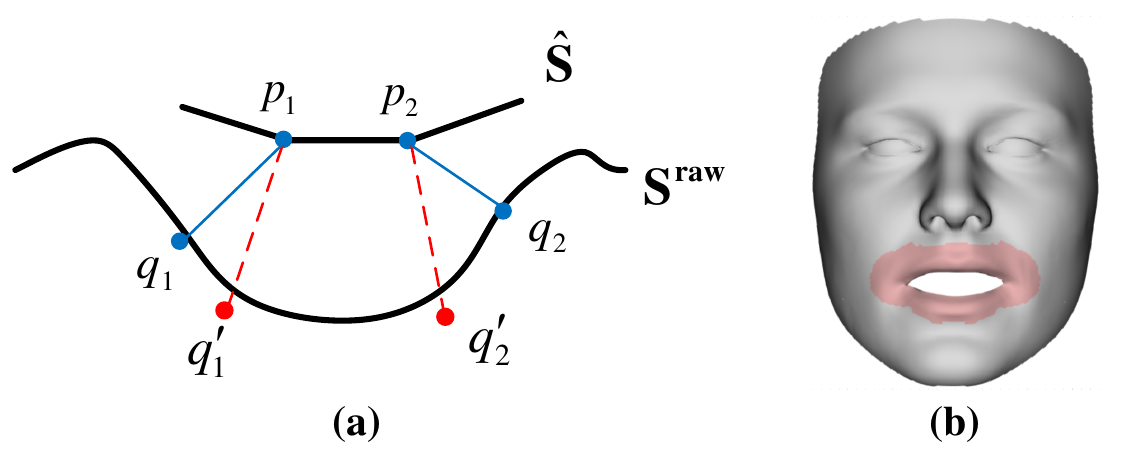}
\vspace{-2mm}
\end{center}
   \caption{\small (a) $q_{i}$ is the closet vertex of $p_{i}$ and ${q}'_{i}$ is computed by the normal ray scheme. (b) the predefined mouth region.}
\label{fig:refine}
\vspace{-4mm}
\end{figure}

\Paragraph{Refine}
In the unsupervised case, the normal loss $\mathcal{L}^{normal}(\hat{\mathbf{n}}, \mathbf{n}^{raw}_{(q)})$ always find the closet vertex $q$ in $\mathbf{S}^{raw}$. 
The disadvantage of this closest vertex scheme is that the counterpart $q_{i}$ is not necessary the true target for correspondence in high-curvature regions (Fig.~\ref{fig:refine}~(a)). Therefore, the loss is not capable of capturing high-frequency details in $\mathbf{S}^{raw}$. 
To remedy this issue, as suggested in~\cite{pan2013establishing}, we consider the normal ray method which computes the closest point of intersection of the normal ray originated from $p_{i}$ with $\mathbf{S}^{raw}$. As shown in Fig.~\ref{fig:refine}~(a), the normal ray in sharp regions would find a better counterpart ${q}'_{i}$. At the early stage of the training process, we use the closet vertex scheme which is computationally more efficient. When the loss gets saturated, we switch to use the normal ray scheme.

For many expressive scans, some tongue points have been recorded when the mouth is open, which are hard to establish correspondence. To address this issue, on top of the losses in Eqn.~\ref{eqn_loss}, we add a mouth region Laplacian regularization loss $\mathcal{L}^{lap}=\left \| L \mathbf{S}_{mouth} \right \|_{2}$ to maintain relative location between neighboring vertices. Here $L$ is the discrete Laplace-Beltrami operator and $\mathbf{S}_{mouth}$ is the mouth region vertex as predefined in Fig.~\ref{fig:refine} (b). See~\cite{Kanazawa_2018_ECCV} for details on Laplacian regularization loss.  



\subsection{Implementation Detail}
\Paragraph{Encoder/Decoder Network} 
We employ the PointNet~\cite{qi2017pointnet} architecture as the base encoder. 
As shown in Fig.~\ref{fig:flowchart}, the encoder takes the $1024$-dim output of PointNet and appends two parallel fully-connected (FC) layers to generate identity and expression latent representations. We set $l_{Id}{=}l_{Exp}{=}512$. 
The decoders are two-layer MLP networks, whose numbers of inputs and outputs are respectively $\{l_{Id}(l_{Exp}), 1024\}$ (ReLU), $\{1024, n{\times} 3\}$. 

\Paragraph{Training Process}
We train our encoder-decoder network in three phases. First, we train the encoder and identity decoder with neutral examples. 
Then, we fix the identity decoder and train the expression decoder with expression examples. 
Finally, the end-to-end joint training is conducted. 
In the first two phases, we start the training with only synthetic data. 
When the loss gets saturated (usually in $10$ epochs), we continue training using a mixture of synthetic and real data for another $10$ epochs. 
We optimize the networks via Adam with an initial learning rate of $0.0001$. 
The learning rate is decreased by half every $5$ epochs. 
We explore the different batch sizes including strategies such as $50\%$ synthetic and $50\%$ real data per batch, and find the optimal batch size to be $1$. 
$\lambda_{1}$ and $\lambda_{2}$ control the influence of regularizations against $ \mathcal{L}^{vt}$. 
They are both set to $1.6{\times}10^{-4}$ in our experiments. We set $\epsilon{=}0.001$ and the weight of $\mathcal{L}^{lap}$ is $0.005$.

\begin{table*}[t]
\renewcommand\arraystretch{0.96}
\centering
\footnotesize
\caption{\small Comparison of the mean and standard deviation of semantic landmark error ($mm$) on BU$3$DFE.}
\vspace{-2mm}
\begin{tabular}{l | c | c | c | c | c |c|c}
\toprule
Face Region & NICP~\cite{amberg2007optimal}& Bolkart~\etal~\cite{bolkart2015groupwise} & Salaza~\etal~\cite{salazar2014fully} & GPMMs~\cite{gerig2018morphable}  & Proposed (\textbf{out}) & Proposed (\textbf{in}) & Relative Impr.\\ 
\hline\hline
Left Eyebrow &      $7.49{\pm}2.04$ & $8.71{\pm}3.32$ & $6.28{\pm}3.30$  & $4.69{\pm}4.64$    &$6.25{\pm}2.58$  &$\bold{4.18{\pm}1.62}$ & $10.9\%$ \\
Right Eyebrow &    $6.92{\pm}2.39$ & $8.62{\pm}3.02$  & $6.75{\pm}3.51$  & $5.35{\pm}4.69$    &$4.57{\pm}3.03$  &$\bold{3.97{\pm}1.70}$ & $25.8\%$ \\
Left Eye &         $3.18{\pm}0.76$ & $3.39{\pm}1.00$    & $3.25{\pm}1.84$  & $3.10{\pm}3.43$  & $2.00{\pm}1.32$  &$\bold{1.72{\pm}0.84}$ & $44.5\%$\\
Right Eye &        $3.49{\pm}0.80$ & $4.33{\pm}1.16$    & $3.81{\pm}2.06$  & $3.33{\pm}3.53$  & $2.88{\pm}1.29$  &$\bold{2.16{\pm}0.82}$ & $35.1\%$\\
Nose &             $5.36{\pm}1.39$ & $5.12{\pm}1.89$    &$3.96{\pm}2.22$  & $3.94{\pm}2.58$  & $4.33{\pm}1.24$  &$\bold{3.56{\pm}1.08}$ & $9.6\%$\\
Mouth &            $5.44{\pm}1.50$ & $5.39{\pm}1.81$   &$5.69{\pm}4.45$  & $\bold{3.66{\pm}3.13}$   &$4.45{\pm}2.02$  &$4.17{\pm}1.70$ & $\text{-} 13.9\%$\\
Chin &              $12.40{\pm}6.15$ & $11.69{\pm}6.39$  &$7.22{\pm}4.73$  & $11.37{\pm}5.85$ &  $7.47{\pm}3.01$  &$\bold{6.80{\pm}3.24}$ & $5.8\%$\\
Left Face &        $12.49{\pm}5.51$ & $15.19{\pm}5.21$  &$18.48{\pm}8.52$ & $12.52{\pm}6.04$ &  $12.10{\pm}4.06$ &$\bold{9.48{\pm}3.42}$ & $24.1\%$\\
Right Face &       $13.04{\pm}5.80$ & $13.77{\pm}5.47$  & $17.36{\pm}9.17$ & $10.76{\pm}5.34$ & $13.17{\pm}4.54$ &$\bold{10.21{\pm}3.07}$ & $5.1\%$\\
\hline
Avg.       &   $7.56{\pm}3.92 $ & $8.49{\pm}4.29$   &  $8.09{\pm}5.75$  & $6.52{\pm}3.86$  & $6.36{\pm}3.92$  &$\bold{5.14{\pm}3.03}$ & $21.2\%$\\
\bottomrule
\end{tabular}
\label{tab:registration_accuracy}
\vspace{-4mm}
\end{table*}

\subsection{Single-Image Shape Inference}
As shown in Fig.~\ref{fig:flowchart}, our identity and expression shape decoders can be used for image-to-shape inference. Specifically, we employ a SOTA face feature extraction network SphereFace~\cite{Liu2017SphereFace} as the base image encoder. This network consists of $20$ convolutional layers and FC layer, and takes the $512$-dim output of the FC layer as the face representation. We append another two parallel FC layers to generate the identity and expression latent representations, respectively. Here, we use the raw scans from the $7$ real databases to render images as our training samples. With the learnt ground truth identity and expression latent codes, we employ a $\mathcal{L}_{1}$ latent loss to fine-tune this image encoder. Since the encoder excels in face feature extraction and latent loss has strong supervision, the encoder is fine-tuned for $100$ epochs with the batch size of $32$.

\section{Experimental Results}
Experiments are conducted to evaluate our method in dense correspondence accuracy, shape and expression representation power, and single-image face reconstruction.

\begin{figure}[t]
\begin{center}
\includegraphics[width=0.88\linewidth]{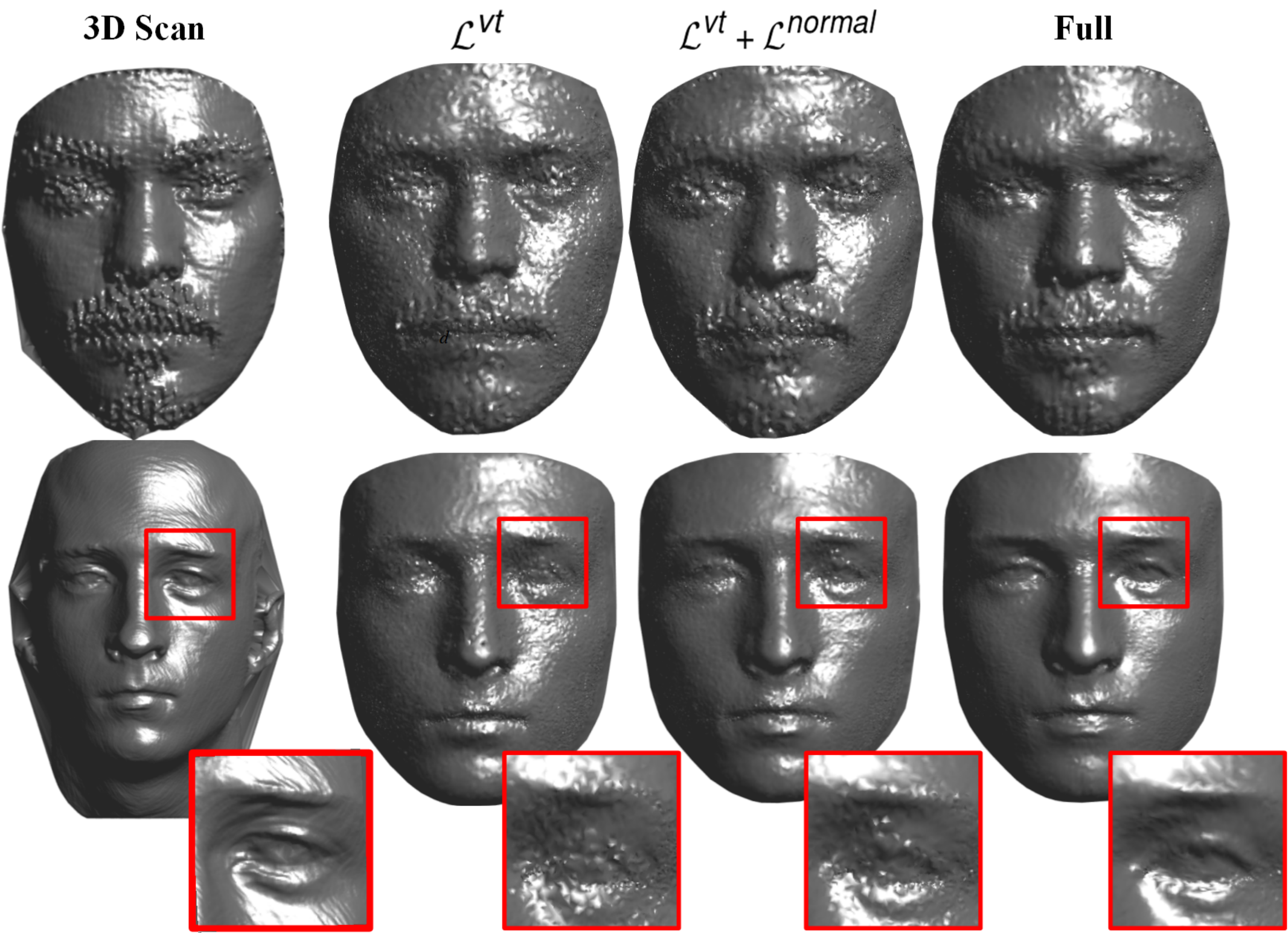}
\vspace{-2mm}
\end{center}
   \caption{\small Qualitative results reflecting the contribution of loss components. The first column is the input scan. Column $2$-$4$ show the reconstructed shapes with different loss combination.}
\label{fig:loss_term}
\vspace{-4mm}
\end{figure}

\Paragraph{Evaluation Metric}
The ideal evaluation metric for $3$D shape analysis is \textbf{\emph{per-vertex error}}. 
However, this metric is not applicable to evaluating real scans due to the absence of dense correspondence ground truth. 
An alternative metric is \textbf{\emph{per-vertex fitting error}}, which has been widely used in $3$D face reconstruction~\cite{liu2018Disentangling} and $3$D face-to-face fitting, \emph{e.g.}, LSFM~\cite{booth2018large}, GPMMs~\cite{luthi2018gaussian}. 
The per-vertex fitting error is the distance between every vertex of the test shape and the nearest-neighbor vertex of the corresponding estimated shape. Generally, the value of this error could be very small due to the nearest-neighbor search. 
Thus, it sometimes can not faithfully reflect the accuracy of dense correspondence.

To better evaluate the correspondence accuracy, prior $3$D face correspondence works~\cite{salazar2014fully, gerig2018morphable, fan2018dense} adopt a \textbf{\emph{semantic landmark error}}. 
With the pre-labeled landmarks on the template face, it is easy to find $p$ $3$D landmarks $\{\hat{l}_{i}\}_{i=1}^{p}$ with the same indexes on the estimated shape.
By comparing with manual annotations $\{l_{i}^{*}\}_{i=1}^{p}$, we can compute the semantic landmark error by $\frac{1}{p}\sum^{p}_{i=1}\left \| l^{*}_{i}-\hat{l}_{i} \right \|$. 
Note that, this error is normally much larger than per-vertex fitting error due to inconsistent and imprecise annotations.  
Tab.~\ref{tab:three_metrics} compares these three evaluation metrics.


\begin{figure}[t]
\begin{center}
\includegraphics[trim=0 0 0 0,clip, width=0.98\linewidth]{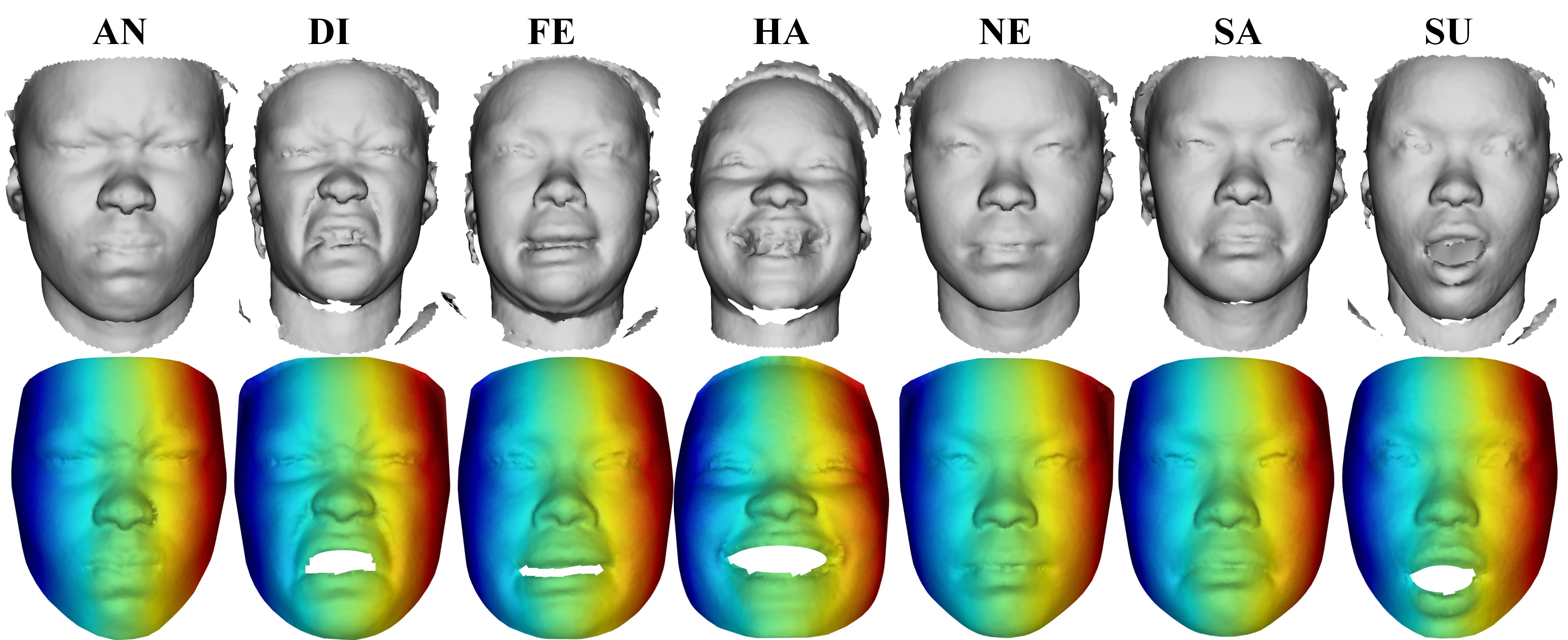}
\vspace{-3mm}
\end{center}
   \caption{\small Raw scans (top) and their reconstructions with color-coded dense correspondences (bottom), for one BU$3$DFE subject in seven expressions: angry (AN), disgust (DI), fear (FE), happy (HA), neutral (NE), sad (SA), and surprise (SU).} 
\label{fig:registration_accuracy}
\vspace{-4mm}
\end{figure}

\subsection{Ablation Study}
We qualitatively evaluate the function of each loss component. As seen in Fig.~\ref{fig:loss_term}, only using vertex loss severely impairs the surface smoothness and local details; adding surface normal loss preserves the high-frequency details. adding edge length term refines the local triangle topology.
These results demonstrate that all the loss components presented in this work contribute to the final performance.


\subsection{Dense Correspondence Accuracy}
We first report the correspondence accuracy on BU$3$DFE database. 
BU$3$DFE contains one neutral and six expression scans with four levels of strength, for each of $100$ subjects.
Following the same setting in~\cite{salazar2014fully} and GPMMs~\cite{gerig2018morphable}, we use all $p=83$ landmarks of all neutral scans and expression scans in the highest level for evaluation. 
Specifically, the landmarks of the estimated shape are compared to the manually annotated landmarks that are provided with BU$3$DFE. 
We compare with four state-of-the-art dense correspondence methods, NICP~\cite{amberg2007optimal}, Bolkart~\etal~\cite{bolkart2015groupwise}, Salazar~\etal~\cite{salazar2014fully}, and GPMMs~\cite{gerig2018morphable}. 
Among them, NICP has been widely used for constructing neutral morphable model such as BFM $2009$ and LSFM. For a fair comparison, we re-implement NICP with extra landmark constraint so that it can establish dense correspondence for expressive $3$D scans. For the other three methods, we report results from their papers. Both Salazar~\etal~\cite{salazar2014fully} and Bolkart~\etal~\cite{bolkart2015groupwise} are multilinear model based $3$D face fitting method. GPMMs~\cite{gerig2018morphable} is a recent Gaussian process registration based method. {\it Note these four baselines do require labeled $3$D landmarks as input, while our method does not.} 


To further evaluate the generalization ability of the proposed method for new scan data, we conduct two series of experiments: (i) training using data from BU$3$DFE database, denoted as Proposed (\textbf{in}), and (ii) training using data outside BU$3$DFE database, denoted as Proposed (\textbf{out}).

As shown in Tab.~\ref{tab:registration_accuracy}, the Proposed (\textbf{in}) setting significantly reduces errors by at least $21.2\%$ w.r.t.~the best baseline.
These results demonstrate the superiority of the proposed method in dense correspondence. 
The error of Proposed (\textbf{out}) setting shows a small increase, but is still lower than the baselines. 
The relatively high semantic landmark error is attributed by the imprecise manual annotations, especially on the semantic ambiguity contour, \emph{i.e.}, Chin, Left Face and Right Face. 
Some example dense correspondence results are shown in Fig.~\ref{fig:registration_accuracy} and \textbf{\emph{Supp.}}. 

\begin{table}[t]
\renewcommand\arraystretch{0.98}
\newcommand{\tabincell}[2]{\begin{tabular}{@{}#1@{}}#2\end{tabular}}
\centering
\scriptsize
\caption{\small Comparison of semantic landmark error (mean+STD in $mm$) on FRGC v2.0. The landmarks are defined in~\cite{fan2018dense}. } 
\vspace{-2mm}
\begin{tabular}{l | c | c | c | c }
\toprule
 Landmark &  \tabincell{c}{Creusot \\ \etal~\cite{creusot2013machine}} &
  \tabincell{c}{ Gilant\\ \etal~\cite{zulqarnain2015shape}} & 
  \tabincell{c}{ Fan\\ \etal~\cite{fan2018dense}}& Proposed\\ 
\hline\hline
ex(L) &    $5.87{\pm}3.11$  & $4.50{\pm}2.97$  &   $2.62{\pm}1.54$ & $\bold{1.79{\pm}1.01}$  \\
en(L) &    $4.31{\pm}2.44$  & $3.12{\pm}2.09$  &   $2.53{\pm}1.66$ & $\bold{1.61{\pm}0.97}$  \\
n     &    $4.20{\pm}2.07$  & $3.63{\pm}2.02$  &   $\bold{2.43{\pm}1.36}$ & $2.69{\pm}1.43$  \\
 ex(R) &    $6.00{\pm}3.03$  & $3.74{\pm}2.79$  &   $2.60{\pm}1.71$ & $\bold{2.00{\pm}0.95}$  \\
 en(R) &    $4.29{\pm}2.03$  & $2.73{\pm}2.14$  &   $2.49{\pm}1.65$ & $\bold{1.82{\pm}0.93}$  \\
 prn   &    $3.35{\pm}2.00$  & $2.68{\pm}1.48$  &   $\bold{2.11{\pm}1.17}$ & $2.36{\pm}1.37$  \\
 Ch(L) &    $5.47{\pm}3.45$  & $5.31{\pm}2.05$  &   $2.93{\pm}2.14$ & $\bold{2.58{\pm}2.61}$  \\
Ch(R) &    $5.64{\pm}3.58$  & $4.38{\pm}2.08$  &   $2.84{\pm}2.17$ & $\bold{2.60{\pm}2.58}$  \\
 ls    &    $4.23{\pm}3.21$  & $3.31{\pm}2.65$  &   $\bold{2.35{\pm}2.86}$ & $2.75{\pm}2.77$  \\
 li    &    $5.46{\pm}3.29$  & $4.02{\pm}3.80$  &   $4.35{\pm}3.93$ & $\bold{4.02{\pm}3.97}$  \\
\hline
Avg.  &    $4.88{\pm}0.91$  & $3.74{\pm}0.83$  &   $2.73{\pm}0.62$ & $\bold{2.42{\pm}0.70}$ \\
\bottomrule
\end{tabular}
\label{tab:registration_accuracy_frgc}
\vspace{-5mm}
\end{table}



We further compare semantic landmark error with the very recent SOTA correspondence method~\cite{fan2018dense}, which is an extension of ICP-based method, on the high-resolution FRGC v2.0 database~\cite{phillips2005overview}. We also compare with two $3$D landmark localization works~\cite{creusot2013machine, zulqarnain2015shape}. Following the same setting in~\cite{fan2018dense}, we compute the mean and standard deviation of $p=10$ landmarks for $4{,}007$ scans. 
The results of baselines are from their papers. 
As shown in Tab.~\ref{tab:registration_accuracy_frgc}, our method improves the SOTA~\cite{fan2018dense} by $11.4\%$, and preserves high-frequency details for high-resolution $3$D models (see \textbf{\emph{Fig.$7$ of Supp.}}). 
The landmark errors are much smaller than BU$3$DFE since the annotations used here are more accurate than BU$3$DFE's. Thanks to the offline training process, our method is two order of magnitude faster than the existing dense correspondence methods: $0.26s$ ($2ms$ with GPU) vs.~$57.48s$ of~\cite{amberg2007optimal} vs.~$164.60s$ of~\cite{fan2018dense}.

\subsection{Representation Power}\label{representation_power}

\begin{figure}[t]
\begin{center}
\includegraphics[width=0.99\linewidth]{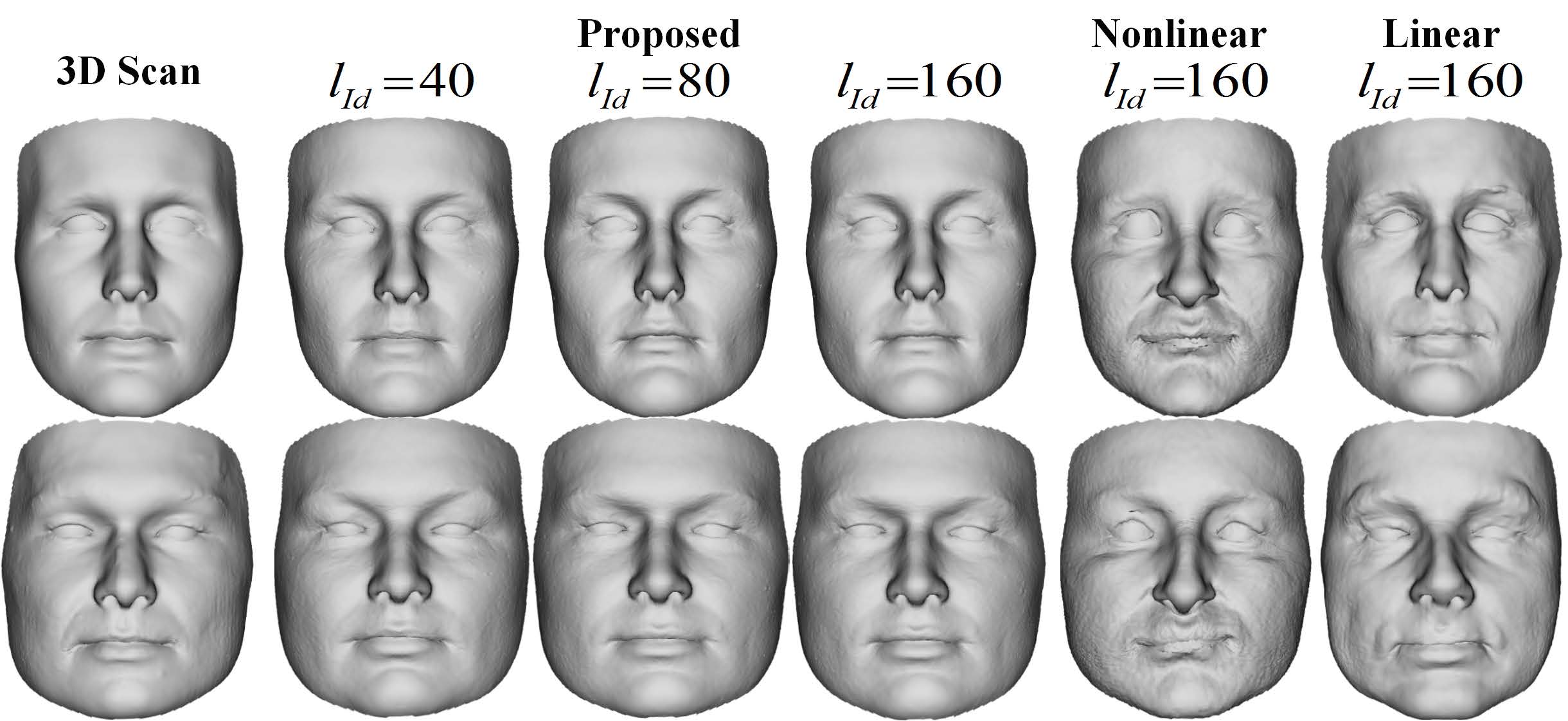}
\vspace{-4mm}
\end{center}
   \caption{\small Shape representation power comparison. Our reconstructions closely match the face shapes and the higher-dim latent spaces can capture more local details.}
\label{fig:shape_representation}
\vspace{-3.5mm}
\end{figure}

\Paragraph{Identity shape} We compare the capabilities of the proposed $3$D face models with linear and nonlinear $3$DMMs on BFM. The BFM database provides $10$ test face scans, which are not included in the training set. As these scans are already established dense correspondence, we use the per-vertex error for evaluation. 
For fair comparison, we train different models with different latent space sizes. 
As shown in Tab.~\ref{tab:shape_representation},  the proposed model has smaller reconstruction error than the linear or nonlinear models. Also, the proposed models are more compact. They can achieve similar performances as linear and nonlinear models whose latent spaces sizes are doubled. Figure~\ref{fig:shape_representation} shows the visual quality of three models' reconstruction. 

\begin{table}[t!]
\renewcommand\arraystretch{0.98}
\footnotesize
\centering
\caption{\small $3$D scan reconstruction comparison (per-vertex error, $mm$). $l_{Id}$ denotes the dimension of latent representation.}
\vspace{-2mm}
\begin{tabular}{l | c | c | c }
\toprule
$l_{Id}$ & $40$ & $80$ & $160$\\ 
\hline\hline
Linear $3$DMM~\cite{paysan20093d}    & $1.669$ & $1.450$ & $1.253$    \\
Nonlinear $3$DMM~\cite{tran2018nonlinear} & $1.440$ & $1.227$ & $1.019$    \\
Proposed         & $\bold{1.258}$ & $\bold{1.107}$ & $\bold{0.946}$    \\
\bottomrule
\end{tabular}
\label{tab:shape_representation}
\vspace{-2mm}
\end{table}

\begin{figure}[t]
\begin{center}
\includegraphics[width=0.99\linewidth]{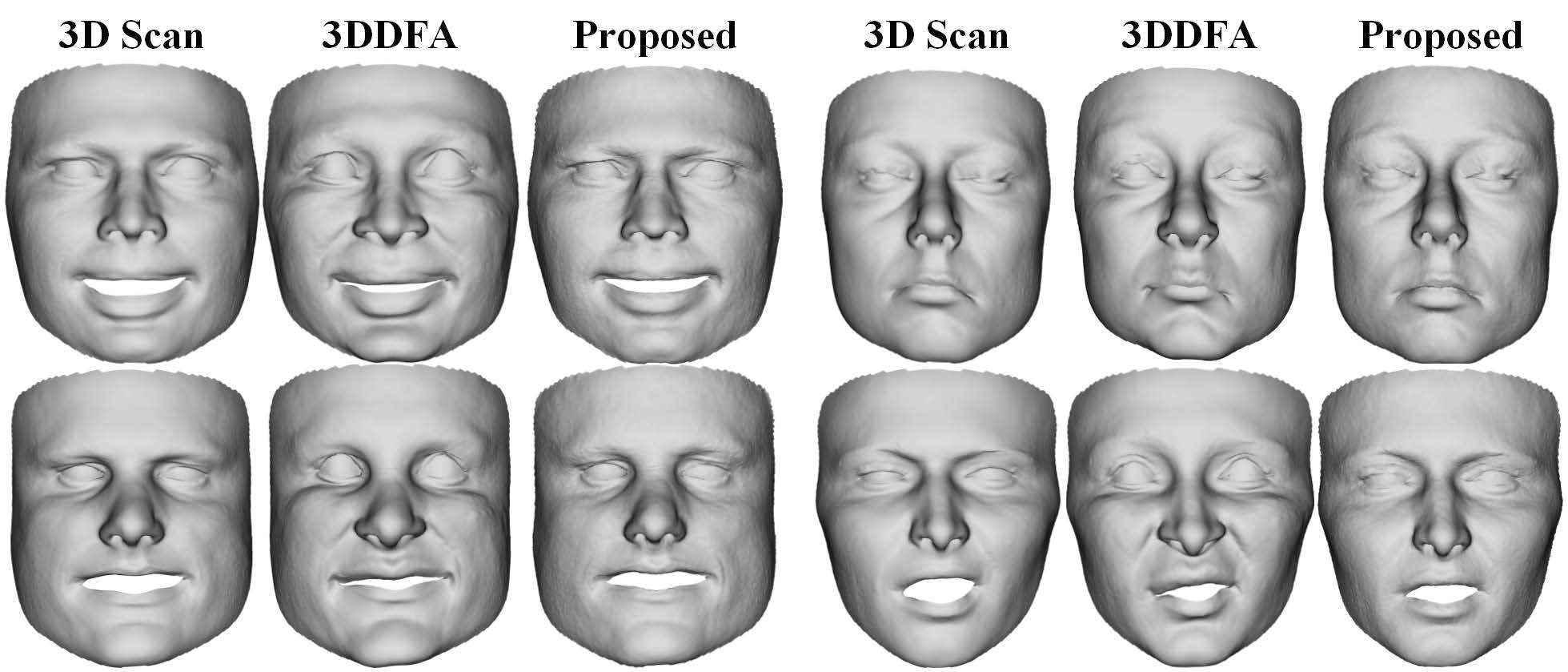}
\vspace{-4mm}
\end{center}
   \caption{\small Expression representation power comparison. Our results better match the expression deformations than $3$DDFA.}
\label{fig:expression_representation}
\vspace{-2mm}
\end{figure}

\begin{table}[t!]
\renewcommand\arraystretch{0.98}
\footnotesize
\centering
\caption{\small Evaluation metric comparison on two databases.}
\vspace{-2mm}
\begin{tabular}{l | c | c  }
\toprule
Metric& BFM  & BU$3$DFE\\ 
\hline\hline
Per-vertex fitting error          & $0.572mm$   &  $1.065mm$   \\
Per-vertex error  &$0.946mm$& $-$  \\
Semantic landmark error   & $1.493mm$ & $5.140mm$   \\
\bottomrule
\end{tabular}
\label{tab:three_metrics}
\vspace{-4mm}
\end{table}

\Paragraph{Expression shape} We compare the expression representation power of the proposed $3$D face models with $3$DDFA expression model~\cite{zhu2015high}, a $29$-dim model originated from FaceWarehouse~\cite{cao2014facewarehouse}. We use a $79$-dim expression model from~\cite{guo2018cnn} to randomly generate an expression difference with Gaussian noise for each BFM test sample. 
Those data are treated as the test set. 
For a fair comparison, we train a model with the same expression latent space size ($l_{Exp}{=}29$).  
Our model has significantly smaller per-vertex error than $3$DDFA: $1.424mm$ vs.~$2.609mm$. Figure~\ref{fig:expression_representation} shows the visual quality of four scans' reconstructions. 

\Paragraph{Shape representation on BU$3$DFE and BFM} 
Table~\ref{tab:three_metrics} compares the shape expressiveness of our model with the three {\it different} metrics. 
Following the setting in Tab.~\ref{tab:shape_representation}, we further calculate the per-vertex fitting error and semantic landmark error ($p=51$) for BFM test samples. 
We also provide the per-vertex fitting error for the BU3DFE reconstructions in Tab~\ref{tab:registration_accuracy}. From Tab.~\ref{tab:three_metrics}, compared to the ideal per-vertex error, semantic landmark error is much larger while per-vertex fitting error is smaller.

\Paragraph{Shape representation on COMA} We further evaluate our shape representation on a large-scale COMA database~\cite{ranjan2018generating}. 
For a fair comparison with FLAME~\cite{li2017learning} and Jiang~\etal~\cite{jiang2019disentangled}, we follow the same setting as~\cite{jiang2019disentangled}, and set our latent vector size as $4$ for identity and $4$ for expression. 
As in Tab.~\ref{tab:coma_result}, our method shows better shape representation compared to SOTA methods. 
While MeshAE~\cite{ranjan2018generating} achieves a smaller error ($1.160mm$), comparing ours with it is not fair, as it has the advantage of encoding $3$D faces into a single vector without decomposing into identity and expression. Also, their mesh convolution requires densely corresponded $3$D scans as input.


\begin{table}[t]
\renewcommand\arraystretch{0.98}
\newcommand{\tabincell}[2]{\begin{tabular}{@{}#1@{}}#2\end{tabular}}
\scriptsize
\centering
\caption{\small Comparison (per-vertex error, $mm$) with state-of-the-art $3$D face modeling methods on COMA database.}
\vspace{-2mm} 
\begin{tabular}{l | c  | c | c  }
\toprule
Sequence & Proposed & Jiang~\etal~\cite{jiang2019disentangled}  & FLAME~\cite{li2017learning} \\ 
\hline\hline
bareteeth       & $\bold{1.609}$  & $1.695$ & $2.002$    \\
cheeks in       & $\bold{1.561}$  & $1.706$ & $2.011$    \\
eyebrow         & $\bold{1.400}$  & $1.475$ & $1.862$     \\
high smile      & $\bold{1.556}$  & $1.714$ & $1.960$      \\
lips back       & $\bold{1.532}$  & $1.752$ & $2.047$     \\
lips up         & $\bold{1.529}$  & $1.747$ & $1.983$      \\
mouth down      & $\bold{1.362}$  & $1.655$ & $2.029$    \\
mouth extreme   & $\bold{1.442}$  & $1.551$ & $2.028$   \\
mouth middle    & $\bold{1.383}$  & $1.757$ & $2.043$   \\
mouth open      & $\bold{1.381}$  & $1.393$ & $1.894$     \\
mouth side      & $\bold{1.502}$  & $1.748$ & $2.090$   \\
mouth up        & $\bold{1.426}$  & $1.528$ & $2.067$     \\\hline
Avg.            & $\bold{1.474}$  & $1.643$ & $1.615$    \\      
\bottomrule
\end{tabular}
\label{tab:coma_result}
\vspace{-2mm}
\end{table}

\subsection{Single-image  $3$D Face Reconstruction}

\begin{figure}[t]
\begin{center}
\includegraphics[width=\linewidth]{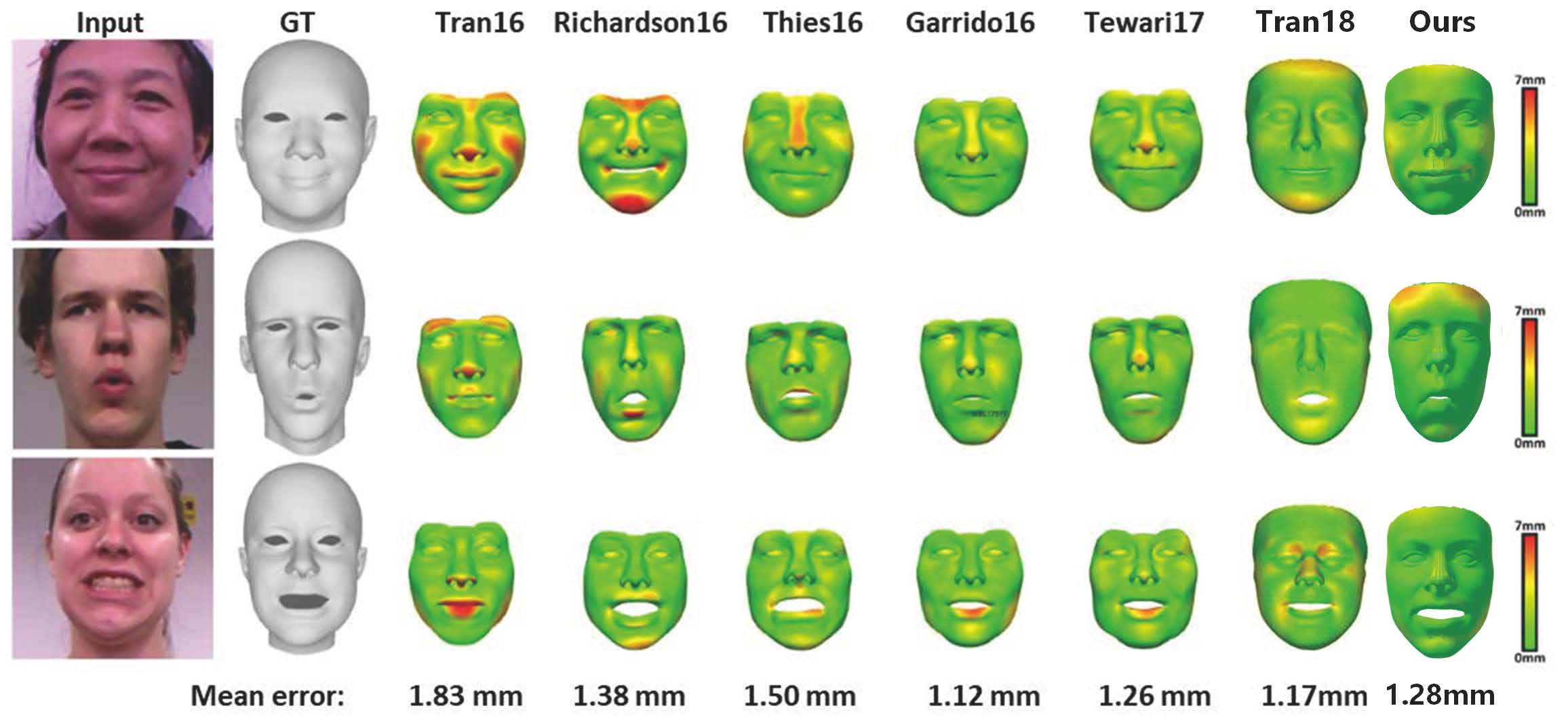}
\vspace{-8mm}
\end{center}
   \caption{\small Quantitative evaluation of single-image $3$D face reconstruction on samples of FaceWarehouse database.}
\label{fig:reconstruction_comparsion}
\vspace{-4mm}
\end{figure}

With the same setting in~\cite{tewari2017mofa}, we quantitatively compare our single-image shape inference with prior works on nine subjects ($180$ images) of the FaceWarehouse database. Visual and quantitative comparisons are shown in Fig.~\ref{fig:reconstruction_comparsion}. We achieve on-par results with nonlinear $3$DMM~\cite{tran2018nonlinear}, Tewari~\cite{tewari2017mofa} and Garrido \emph{et al.}~\cite{garrido2016reconstruction}, while surpassing all other CNN-based regression methods~\cite{tran2016regressing, richardson2016learning}.


%
%

\section{Conclusions}
This paper proposes an innovative encoder-decoder to jointly learn a robust and expressive face model from a diverse set of raw $3$D scan databases and establish dense correspondence among all scans. By using a mixture of synthetic and real $3$D scan data with an effective weakly-supervised learning-based approach, our network can preserve high-frequency details of $3$D scans. The comprehensive experimental results show that the proposed method can effectively establish point-to-point dense correspondence, achieve more representation power in identity and expression, and is applicable to $3$D face reconstruction. 

\Paragraph{Acknowledgment } Research was sponsored by the Army Research Office and was accomplished under Grant Number W911NF-18-1-0330. The views and conclusions contained in this document are those of the authors and should not be interpreted as representing the official policies, either expressed or implied, of the Army Research Office or the U.S. Government. The U.S. Government is authorized to reproduce and distribute reprints for Government purposes notwithstanding any copyright notation herein.

{\small
\bibliographystyle{ieee_fullname}
\bibliography{egbib}
}

\end{document}